\documentclass[10pt,twocolumn,letterpaper]{article}
\usepackage{iccv}
\usepackage{times}
\usepackage{epsfig}
\usepackage{graphicx}
\usepackage{amsmath}
\usepackage{amssymb}
\usepackage{bm}

\usepackage{array}
\usepackage{multirow}
\usepackage[misc]{ifsym}
\usepackage[accsupp]{axessibility}

\newcolumntype{P}[1]{>{\centering\arraybackslash}p{#1}}

\usepackage[breaklinks=true,bookmarks=false]{hyperref}

\newcounter{alphasect}
\def\alphainsection{0}

\let\oldsection=\section
\def\section{%
  \ifnum\alphainsection=1%
    \addtocounter{alphasect}{1}
  \fi%
\oldsection}%

\renewcommand\thesection{%
  \ifnum\alphainsection=1%
    \Alph{alphasect}
  \else%
    \arabic{section}
  \fi%
}%

\newenvironment{alphasection}{%
  \ifnum\alphainsection=1%
    \errhelp={Let other blocks end at the beginning of the next block.}
    \errmessage{Nested Alpha section not allowed}
  \fi%
  \setcounter{alphasect}{0}
  \def\alphainsection{1}
}{%
  \setcounter{alphasect}{0}
  \def\alphainsection{0}
}%

\makeatletter
\newcommand\xleftrightarrow[2][]{%
  \ext@arrow 9999{\longleftrightarrowfill@}{#1}{#2}}
\newcommand\longleftrightarrowfill@{%
  \arrowfill@\leftarrow\relbar\rightarrow}
\makeatother

\iccvfinalcopy 

\def\iccvPaperID{1996} 

\ificcvfinal\fi

\begin{document}

\title{PIRenderer: Controllable Portrait Image Generation via  Semantic Neural Rendering}
\author{Yurui Ren$^1$~~~{Ge Li \footnotesize{\Letter}}$^1$~~~Yuanqi Chen$^1$~~~Thomas H. Li$^{2,1}$~~~Shan Liu$^{3}$~~~\\
$^1$School of Electronics and Computer Engineering, Peking University~~~\\
$^2$Advanced Institute of Information Technology, Peking University~~~$^3$Tencent America~~~~~~~~~\\
{\tt\small \{yrren,geli,cyq373\}@pku.edu.cn~~~tli@aiit.org.cn~~~shanl@tencent.com}
}



\maketitle
\ificcvfinal\thispagestyle{empty}\fi

\begin{abstract}
Generating portrait images by controlling the motions of existing faces is an important task of great consequence to social media industries. For easy use and intuitive control, semantically meaningful and fully disentangled parameters should be used as modifications. 
However, many existing techniques do not provide such fine-grained controls or use indirect editing methods i.e. mimic motions of other individuals. 
In this paper, a Portrait Image Neural Renderer (PIRenderer) is proposed to control the face motions with the parameters of three-dimensional morphable face models (3DMMs). 
The proposed model can generate photo-realistic portrait images with accurate movements according to intuitive modifications. 
Experiments on both direct and indirect editing tasks demonstrate the superiority of this model. 
Meanwhile, we further extend this model to tackle the audio-driven facial reenactment task by extracting sequential motions from audio inputs. We show that our model can generate coherent videos with convincing movements from only a single reference image and a driving audio stream. Our source code is available at \url{https://github.com/RenYurui/PIRender}.
\end{abstract}

\begin{figure}[t]
\begin{center}
\includegraphics[width=0.9\linewidth]{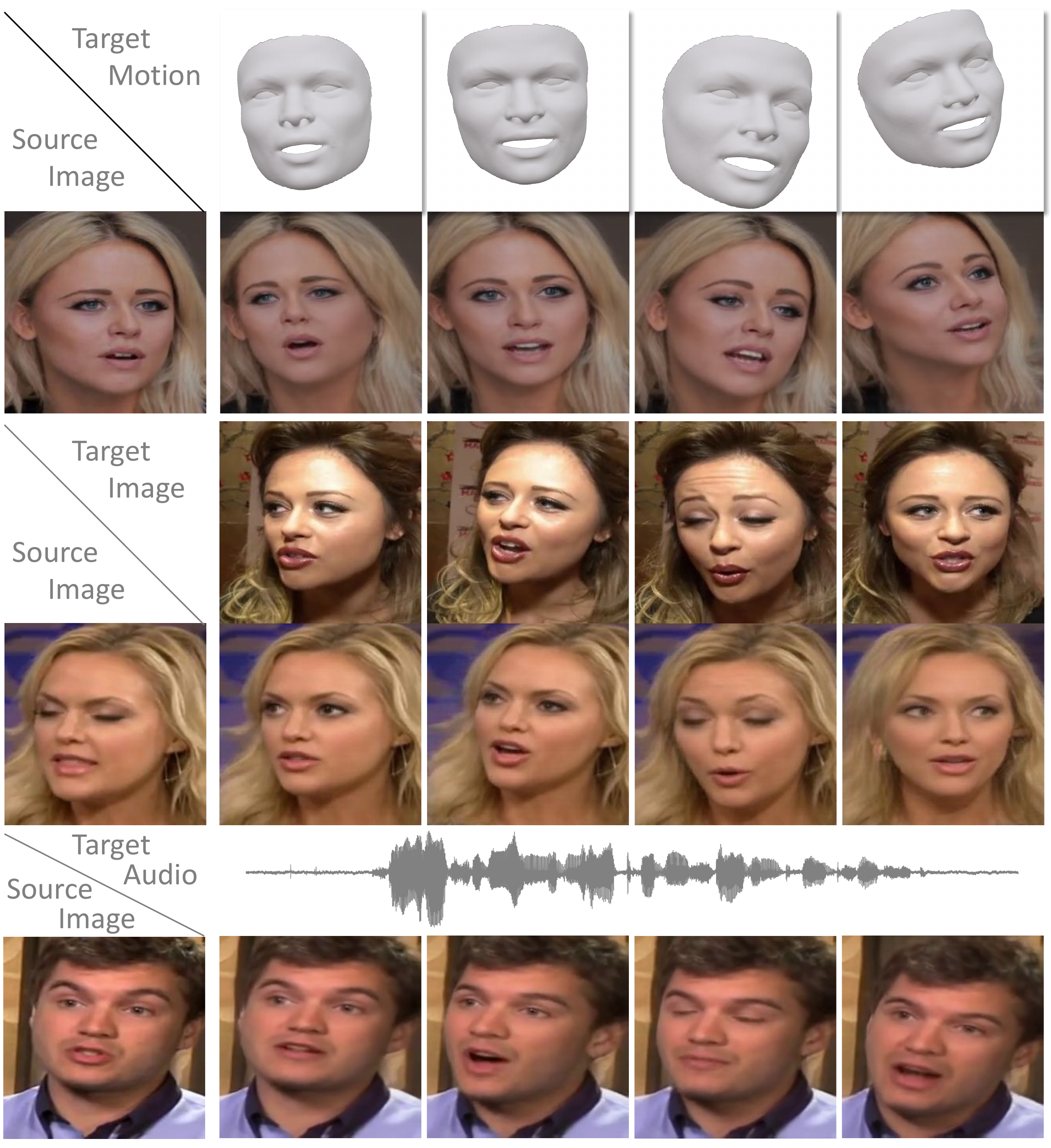}
\end{center}
   \caption{Example results produced by our PIRenderer. This model can generate photo-realistic portrait images according to the user-specified motions (top), motions of another individual (middle), and motions generated from audios (bottom).}

 
\label{fig:title_task}
\end{figure}
\section{Introduction}
Portrait images are one of the most important photographic depictions that are widely used in daily life.
Being able to edit portrait images by intuitively controlling the poses and expressions of given faces (see Fig.~\ref{fig:title_task}) is an important task with a large variety of applications in the fields of virtual reality, film production, and next-generation communication.
However, enabling such editing is extremely challenging since it requires the algorithm to perceive reliable 3D geometric shapes of given faces.
Meanwhile, the acuteness of the human visual system towards portrait images requires the algorithm to generate photo-realistic faces and backgrounds, which makes the task even harder.

Recently, advances in Generative Adversarial Networks (GANs)~\cite{goodfellow2014generative} have made tremendous progress in synthesizing realistic faces~\cite{brock2018large,karras2019style,choi2018stargan}. Some methods~\cite{bansal2018recycle,wang2018video} driven from GANs tackle this task with image translation techniques, where the goal is to train a model such that the conditional distribution of the generated images given input instructions (\eg edges) resembles that of real images. Some follow-up algorithms achieve better generalization by proposing efficient deformation modules~\cite{wiles2018x2face,zakharov2019few,siarohin2019animating,chen2020puppeteergan,ren2020deep,siarohin2019first} or injecting source neural textures to targets~\cite{wang2019few,burkov2020neural}.
However, the vast majority of existing methods use indirect and subject-specific motion descriptors such as edges~\cite{wang2018video,wang2019few,ren2020deep}, semantic segmentation~\cite{chen2020puppeteergan}, or keypoints~\cite{siarohin2019animating,siarohin2019first,ren2020deep} to describe the target motions. 
Although these descriptors with 2D spatial information can benefit the generation of target images, they hinder the model ability to edit portraits in an intuitive manner.

To achieve intuitive control, motion descriptors should be semantically meaningful, which requires facial expressions, head rotations, and translations to be expressed as fully disentangled variables.
Parametric face modeling methods~\cite{blanz1999morphable,paysan20093d} provide powerful tools for describing 3D faces with semantic parameters. These methods allow controlling 3D meshes over parameters such as shape, expressions, \etc.
Incorporating priors of these techniques, one can expect to control the generation of photo-realistic portrait images akin to the processing of graph rendering.

\begin{figure*}[t]
\begin{center}
\includegraphics[width=1\linewidth]{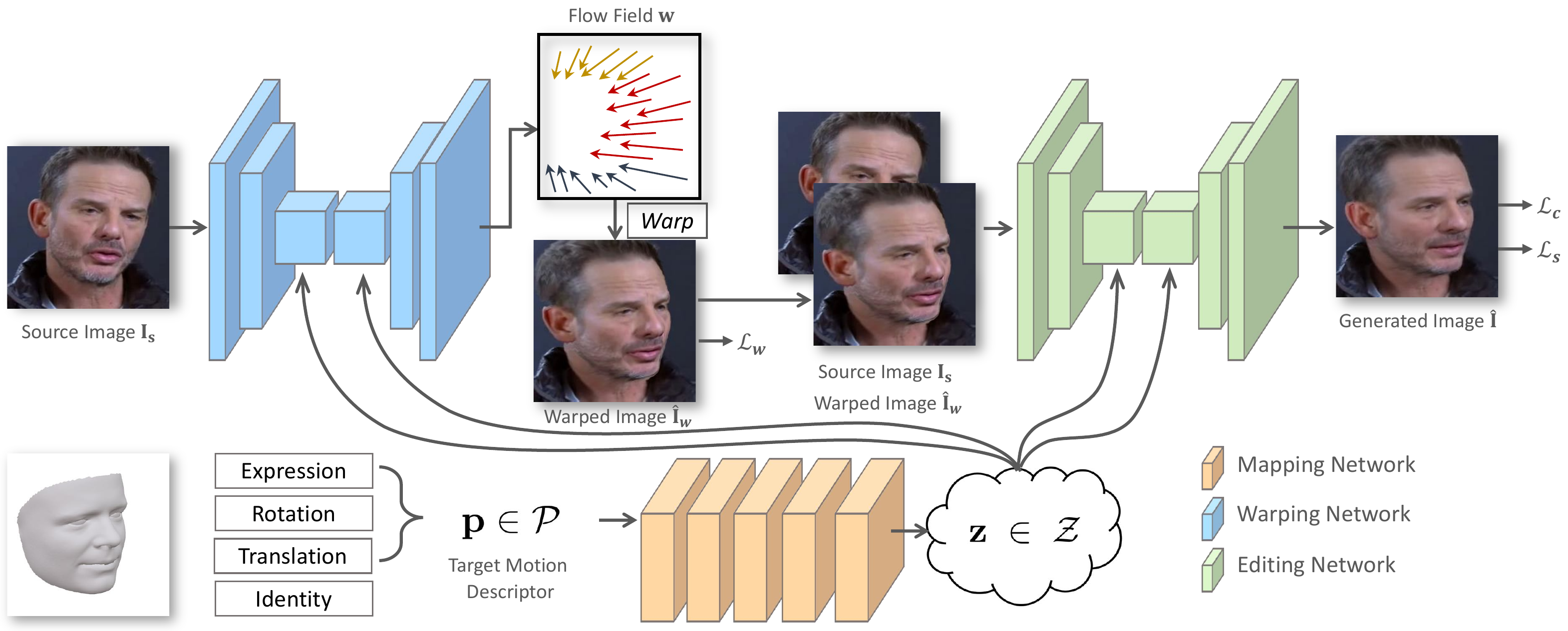}
\end{center}
   \caption{Overview of the PIRenderer. Our model first maps the input motion descriptor $\mathbf{p}$ to a latent vector $\mathbf{z}$. Then a coarse image is generated by the warping network. Finally, the editing network is responsible for generating the final image by editing the coarse result. }
 
\label{fig:OurModel}
\end{figure*}

In this paper, a neural rendering model PIRenderer is proposed.
Given a source portrait image and target 3DMM parameters, our model generates photo-realistic results with accurate motions. 
As shown in Fig.~\ref{fig:OurModel}, the proposed model is divided into three parts:
the Mapping Network, the Warping Network, and the Editing Network. 
The mapping network produces latent vectors from motion descriptors.
Instructed by the vectors, the warping network estimates deformations between sources and desired targets and generates coarse results by warping sources with the estimated deformations.
The editing network generates the final images from the coarse images.
The superiority and versatility of our model are demonstrated in the experiments. 
We show that our model not only enables intuitive image control by editing target images with user-specified motions but also generates realistic results in the indirect portrait editing task where the goal is to mimic the motions of another individual.
Additionally, we show the potential of our model as an efficient neural renderer by further extending it to tackle the audio-driven facial reenactment task. Thanks to the high-level fully disentangled parameterizations, we can extract convincing motions from ``weak'' control audios. Experiments show that our model generates various and vivid motions from an audio stream and transfers the motions into realistic videos of arbitrary target persons. The main contributions are summarized as:
\vspace{-2.5mm}
\begin{itemize}
  \setlength\itemsep{-0.9mm}
  \item We proposed a portrait image generation model PIRenderer, which enables intuitive photo-real editing of facial expressions, head rotations, and translations. 
  \item The proposed model can be used to tackle the indirect image editing task which requires 
  imitating
  other individuals. Thanks to our disentangled modifications and efficient neural renderer, we can extract subject-agnostic motions and generate realistic videos.
  \item Additional extension on audio-driven facial reenactment demonstrates the potential of the proposed model as an efficient face renderer. Our model generates various and vivid videos from only a single portrait image and a driving audio stream.

\end{itemize}

\section{Related Work}
We define portrait editing as the processing of modifying the facial expressions, head rotations, and translations of a given image or video. 
The existing methods are divided into the following three parts according to the controlling signals.

\vspace{1mm}
\noindent
\textbf{Portrait Editing via Semantic Parameterization.~} 
Editing portrait images using a semantic control space akin to computer animation controls can provide users with intuitive controls.
Some model-based methods incorporate the rendering images of 3DMMs and edit portrait images by modifying the expression~\cite{thies2019deferred,thies2016face2face} or pose~\cite{kim2018deep} parameters. These methods achieve impressive results, but they are subject-specific methods, which means that they cannot be applied to arbitrary persons.
X2Face~\cite{wiles2018x2face} achieves better generalization by employing the warping operation to spatially transform source textures. However, it can only edit the pose of source portraits. 
Recently, generative adversarial networks such as StyleGAN~\cite{karras2019style} were trained to synthesis realistic face images. Capitalizing on the advantages of these techniques, StyleRig~\cite{tewari2020stylerig} describes a method to control StyleGAN via a 3DMM. However, because it is trained to map latent variables of StyleGAN, it does not support real-world image editing. PIE~\cite{tewari2020pie} deals with this shortcoming by proposing an optimization approach to calculate the corresponding embeddings of real images. 
However, the iterative optimization operation is needed for each input image, which reduces the efficiency of the algorithm.

\vspace{1mm}
\noindent
\textbf{Portrait Editing via Motion Imitation.~} 
Instead of describing target motions with semantic parameters, many models~\cite{wiles2018x2face,bansal2018recycle,wang2018video,siarohin2019animating,wang2019few,ren2020deep,siarohin2019first,ren2020deepspatial,zakharov2020fast} are trained to mimic the motions of another individual. Several methods~\cite{bansal2018recycle,wu2018reenactgan,wang2018video} derived from generative adversarial networks~\cite{goodfellow2014generative} have been proposed to tackle this task by modeling it as the video counterpart of image-to-image translation~\cite{isola2017image,zhu2017unpaired}. These approaches require training on hours of labeled videos of a single person to perform reenactment and therefore have to be retrained for each individual. Some follow-up methods achieve better generalization by proposing efficient spatial transformation modules~\cite{siarohin2019animating,siarohin2019first,ren2020deep,wang2020one} or injecting source neural textures to targets~\cite{zakharov2019few,burkov2020neural,zakharov2020fast}. These methods can generate realistic images for arbitrary persons. However, they rely on subject-specific (\eg landmarks, edges, parsing maps) or motion-entangled (\eg sparse keypoints) descriptors, which leaves them lacking the ability to intuitively edit source portraits.

\vspace{1mm}
\noindent
\textbf{Portrait Editing via Audio.~} 
Editing portrait images with audios requires generating coherent videos with convincing motions from audio streams and source images. Some methods~\cite{chung2017you,vougioukas2018end,song2018talking,zhou2019talking} approach this task by directly modeling the relationships between audio signals and images. 
However, since audio inputs cannot fully determine the head poses of targets, these methods are trained to generate videos with a fixed head pose and therefore do not support full head editing.
To avoid interference from the irrelevant factors in real imagery, some models propose to first map audio inputs to intermediate results (\eg landmarks~\cite{suwajanakorn2017synthesizing,chen2019hierarchical}, 3DMMs~\cite{thies2020neural,yi2020audio,wen2020photorealistic}), and then generate the final images. Incorporating the priors of 3DMMs, papers~\cite{suwajanakorn2017synthesizing,thies2020neural,wen2020photorealistic,yi2020audio} generate targets by synthesizing realistic textures for the rendered images and blending the results with source images. 
However, these methods require video input and do not work on single images. Meanwhile, they are either subject-specific models~\cite{suwajanakorn2017synthesizing} or require further fine-tuning before applied to arbitrary individuals~\cite{thies2020neural,yi2020audio}.
Paper~\cite{wiles2018x2face} supports full head editing from single images by mapping audio signals to latent variables. However, it does not model the variation of possible motions.
Our model can generate various motions for an audio stream and transfer these motions into realistic videos of arbitrary target persons.

\section{Our Approach}

We propose a semantic neural rendering model PIRenderer for controllable portrait image generation. Instructed by the target motion descriptors $\mathbf{p}$, our model can generate photo-realistic portrait image $\hat{\mathbf{I}}$ by modifying the face expressions, head poses, and translations of a given image $\mathbf{I}_s$ while maintaining other source attributes such as identity, illumination, and background. 
In the following, we first provide the details about the motion descriptors (Sec. \ref{Semantic_Controlling_Parameters}). Then, we introduce our PIRenderer for portrait image semantic control (Sec. \ref{Network_Architecture}). 
Finally, we extend our model to tackle a more complex sequential generation task audio-driven facial reenactment (Sec. \ref{Audio-driven_Reenactment}).

\subsection{Target Motion Descriptor}
\label{Semantic_Controlling_Parameters}
To achieve intuitive and fine-grained editing, semantically meaningful controls should be provided. In this paper, we employ a subset of 3DMM parameters as the motion descriptor.
With a 3DMM, the 3D shape $\mathbf{S}$ of a face is parameterized as
\begin{equation}
    \mathbf{S} = \bar{\mathbf{S}} + \bm{\alpha}\mathbf{B}_{id} + \bm{\beta}\mathbf{B}_{exp}
\end{equation}
where $\bar{\mathbf{S}}$ is the average face shape, $\mathbf{B}_{id}$ and $\mathbf{B}_{exp}$ are the bases of identity and expression computed via Principal Component Analysis (PCA) based on 200 scans of human faces~\cite{paysan20093d}. 
Coefficients $\bm{\alpha} \in \mathbb{R}^{80}$ and $\bm{\beta} \in \mathbb{R}^{64}$ describe the facial shape and expression respectively. The head rotation and translation are expressed as $\mathbf{R} \in SO(3)$ and $\mathbf{t} \in \mathbb{R}^3$. With parameter set $\mathbf{p}_i \equiv \{\bm{\beta}_i, \mathbf{R}_i, \mathbf{t}_i \}$, the desired motions of face $i$ can be clearly expressed.

An off-the-shelf 3D face reconstruction model~\cite{deng2019accurate} is employed for extracting the corresponding 3DMM coefficients
from real-world portrait images $\mathbf{I}_t$ for training and evaluation.
However, coefficient extraction creates the problem of estimation errors.
Although the 3D face reconstruction method produces relatively accurate results, errors and noise are inevitable.
Mismatches between the extracted motions $\mathbf{p}$ and the real motions of targets $\mathbf{I}_t$ lead to performance decline and incoherent results (see supplementary materials).
To alleviate this problem, the coefficients of a window with continuous frames are used as the motion descriptor of the center frame. 
Thus, the network can be expected to avoid errors by extracting relationships between adjacent frames. Therefore, the motion descriptor is defined as $\mathbf{p} \equiv \mathbf{p}_{i-k:i+k} \equiv \{ \bm{\beta}_i, \mathbf{R}_i, \mathbf{t}_i, ..., \bm{\beta}_{i\pm k}, \mathbf{R}_{i\pm k}, \mathbf{t}_{i \pm k \}}$, where $k$ is the radius of the window.

\subsection{PIRenderer for Semantic Control}

\label{Network_Architecture}
Given a source portrait image $\mathbf{I}_s$ and target motion descriptor $\mathbf{p}$, the PIRenderer generates a portrait image $\hat{\mathbf{I}}$ with accurate target motions while maintaining the other source information such as identity, illumination, and background. Fig.~\ref{fig:OurModel} shows an overview of the propositioned architecture. The network combines several components that fulfill specific tasks.

\vspace{2mm}
\noindent
\textbf{The Mapping Network.~} We employ a mapping network $f_m:\mathcal{P} \rightarrow \mathcal{Z}$ to produce latent vectors $\mathbf{z} \in \mathcal{Z}$ from the motion descriptors $\mathbf{p} \in \mathcal{P}$. 
\begin{equation}
  \mathbf{z} = f_m(\mathbf{p})
\end{equation}
The learned latent vectors $\mathbf{z}$ are further transformed by affine transformations to generate $\mathbf{y}=(\mathbf{y}_s, \mathbf{y}_b)$ that control adaptive instance normalization (AdaIN)~\cite{huang2017arbitrary} operations.
The AdaIN operations are responsible for injecting the motions described by $\mathbf{z}$ to the warping and the editing network. It is defined as
\begin{equation}
  \textup{AdaIN}(\mathbf{x}_i, \mathbf{y}) = \mathbf{y}_{s,i} \frac{\mathbf{x}_i-\mu(\mathbf{x}_i)}{\sigma(\mathbf{x}_i)} + \mathbf{y}_{b,i}
\end{equation}
where $\mu(\cdot)$ and $\sigma(\cdot)$ represent the average and variance operations respectively. Each feature map $\mathbf{x}_i$ is first normalized and then scaled and biased using the corresponding scalar components of $\mathbf{y}$.

\vspace{2mm}
\noindent
\textbf{The Warping Network.~} Convolutional Neural Networks lack the capability to spatially transform the inputs in an efficient manner. To better preserve vivid source textures and achieve better generalization, we use a warping network $g_w$ to spatially transform vital information of the source images $\mathbf{I}_s$. 
The warping network is expected to perceive accurate 3D geometric shapes of source faces and estimates the deformations between input sources and desired targets. 
It takes the source image $\mathbf{I}_s$ and the latent vector $\mathbf{z}$ as inputs and generates the flow field $\mathbf{w}$ containing the coordinate offsets specifying which positions in the sources could be sampled to generate the targets.
\begin{equation}
  \mathbf{w} = g_w(\mathbf{I}_s, \mathbf{z})
\end{equation}
The network is designed with an auto-encoder architecture. AdaIN operation is used after each convolution layer to inject the motions described by $\mathbf{z}$. 
After papers~\cite{teed2020raft,siarohin2019first,ren2020deep,jonschkowski2020matters}, we do not estimate full resolution flow fields. The output flow is at the $1/4$ resolution of the input image. During training and evaluation, we upsample the predicted flow fields to match the resolution. 
After obtaining the flow fields $\mathbf{w}$, the coarse results $\hat{\mathbf{I}}_w$ can be calculated by the equation $\hat{\mathbf{I}}_w=\mathbf{w}(\mathbf{I}_s)$ where we warp the source image $\mathbf{I}_s$ with $\mathbf{w}$.

We constrain the warping network for generating accurate flow fields with a warping loss $\mathcal{L}_w$. As the ground-truth flow fields are not available in this task, we calculate the reconstruction error between the warped image $\hat{\mathbf{I}}_w$ and the target image $\mathbf{I}_t$. The warping loss is based on the perceptual loss of Johnson \etal~\cite{johnson2016perceptual} which calculates the $\ell_1$ distance between activation maps of the pre-trained VGG-19 network.
\begin{equation}
  \mathcal{L}_w = \sum_i\left\lVert \phi_i(\mathbf{I}_t) - \phi_i(\hat{\mathbf{I}}_w)\right\lVert_1
\end{equation}
where $\phi_i$ is the activation map of the $i$-th layer of the VGG-19 network. Similar to~\cite{siarohin2019first}, we calculate the perceptual loss on a number of resolutions by applying pyramid down-sampling on $\mathbf{I}_t$ and $\hat{\mathbf{I}}_w$. 

\vspace{2mm}
\noindent
\textbf{The Editing Network.~} 
Although the warping network is efficient at spatially transforming the source images, it is limited by the inability to generate contents that do not exist in the sources. Meanwhile, artifacts introduced by the warping operation will lead to performance decline. Therefore, an editing network $g_e$ is designed to modify the warped coarse results $\hat{\mathbf{I}}_w$.
The editing network takes $\hat{\mathbf{I}}_w$, $\mathbf{I}_s$, and $\mathbf{z}$ as inputs and generates the final prediction $\hat{\mathbf{I}}$.
\begin{equation}
  \hat{\mathbf{I}} = g_e(\hat{\mathbf{I}}_w, \mathbf{I}_s, \mathbf{z})
\end{equation}
The editing network is designed using a similar architecture to that of the warping network. The AdaIN operation is also employed to inject the latent vector $\mathbf{z}$. We use skip-connect layers in the editing network to preserve the source textures. 
More details of our network architecture can be found in the supplementary materials.

Our editing network is trained with a reconstruction loss $\mathcal{L}_c$ and a style loss $\mathcal{L}_s$. The Reconstruction Loss is used to reduce the errors between the final predictions $\hat{\mathbf{I}}$ and the ground-truth targets $\mathbf{I}_t$. It is designed with similar architecture as that of the warping loss $\mathcal{L}_w$. 
\begin{equation}
  \mathcal{L}_c = \sum_i\left\lVert \phi_i(\mathbf{I}_t) - \phi_i(\hat{\mathbf{I}})\right\lVert_1
\end{equation}
Similarly, pyramid down-sampling is applied to $\mathbf{I}_t$ and $\hat{\mathbf{I}}$ to calculate the loss on multi-scale images.
The style loss $\mathcal{L}_s$ calculates the statistic error between the VGG-19 activation maps as
\begin{equation}
    \mathcal{L}_s = \sum_j\left\lVert G_j^\phi(\mathbf{I}_t) - G^\phi_j(\hat{\mathbf{I}})\right\lVert_1
\end{equation}
where $G_j^\phi$ is the Gram matrix constructed from activation maps $\phi_j$.
The final loss of the proposed model is a summation of the above losses.
\begin{equation}
  \mathcal{L} = \lambda_w \mathcal{L}_w + \lambda_c \mathcal{L}_c + \lambda_s \mathcal{L}_s
\end{equation}
In the experiments, we set $\lambda_w=2.5$, $\lambda_c=4$, and $\lambda_s=1000$.

\begin{figure}[t]
\begin{center}
\includegraphics[width=1\linewidth]{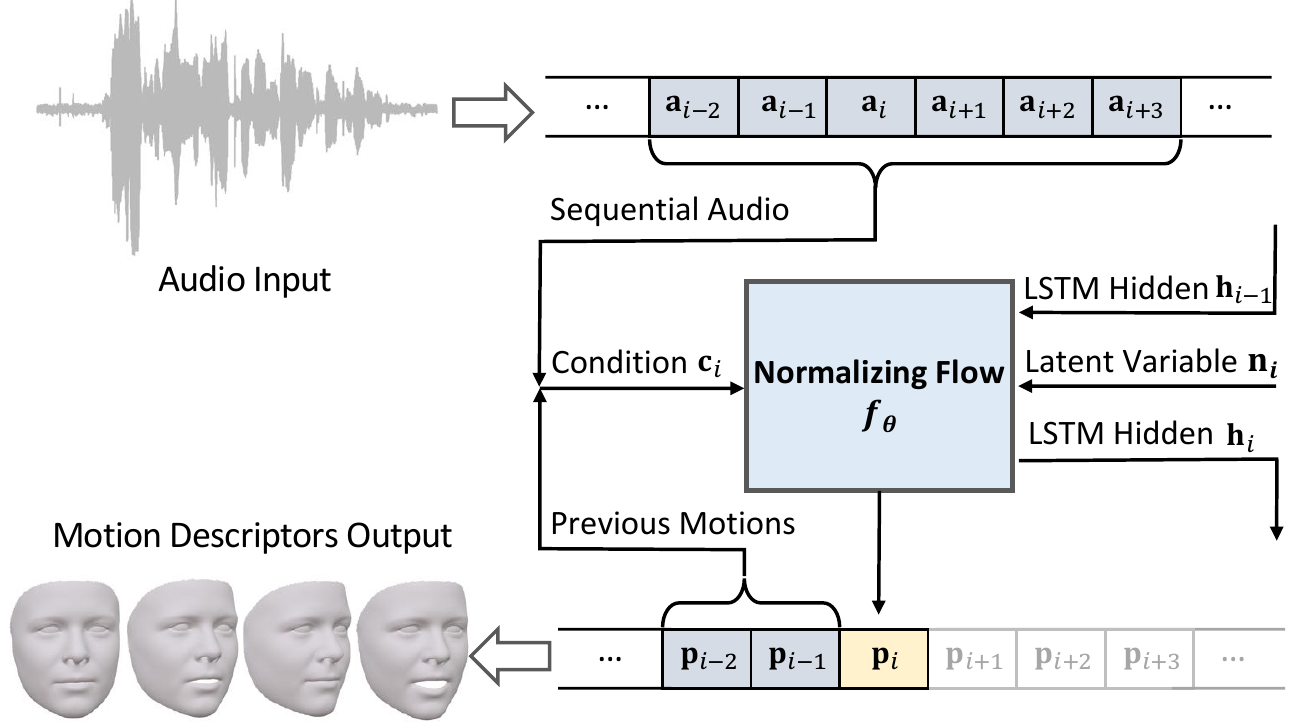}
\end{center}
   \caption{Overview of the audio-driven reenactment model. The previously generated motions $\mathbf{p}_{i-k:i-1}$ and the audios $\mathbf{a}_{i-k:i+\tau}$ are used as the conditional information to generate current motion $\mathbf{p}_i$. 
   Our normalizing flow $f_\theta$ is used to map the latent variable $\mathbf{n}_i \in \mathcal{N}$ to the motion descriptor $\mathbf{p}_i \in \mathcal{P}$.}
 
\label{fig:norm_flow}
\end{figure}

\subsection{Extension on Audio-driven Reenactment}
\label{Audio-driven_Reenactment}
In this subsection, we further extend the PIRenderer to tackle the audio-driven facial reenactment task which generates videos with convincing expressions and poses from driving audios and source images.
This task requires one to model the relationships between audios and face motions. 
However, directly mapping audio signals to photo-realistic imagery or other low-level motion descriptors (\eg edges, landmarks) is challenging. Motion-irrelevant factors such as identity, and illumination will interfere with the model.
Therefore, employing semantically meaningful parameterizations like 3DMMs as the intermediate results can significantly ease the task. 
Consequently, we further improve our model by including an additional mapping function $f_\theta$ to produce sequential 3DMM coefficients from audios.

A fundamental challenge of designing $f_\theta$ is to handle the large variety of possible outputs. 
Training it to generate deterministic motions will limit the network to produce artifacts as it is constrained to predict the mean motions of all possible results.
For this reason, outlining $f_\theta$ as a stochastic model that can generate various motions from a single audio stream assists in reducing this issue. 
\textit{Normalizing flow}~\cite{dinh2014nice,rezende2015variational,kingma2018glow} is employed to design this model. 
The core idea of normalizing flow is to train an invertible and differentiable nonlinear mapping function that maps samples from a simple distribution to a more complex distribution.
In the conditional setting, the model can be trained to map sequential motion-condition pairs $(\mathbf{p}, \mathbf{c}) \equiv (\mathbf{p}_{1:t}, \mathbf{c}_{1:t})$ to latent variables $\mathbf{n}$ with 
\begin{equation}
\mathbf{n} = f_\theta^{-1}(\mathbf{p}, \mathbf{c})  
\end{equation}
By postulating a simple distribution $p_\mathbf{n}(\mathbf{n})$ (\eg a Gaussian $\mathcal{N}$), the conditional distribution $p_{\mathbf{p}|\mathbf{c}}(\mathbf{p} | \mathbf{c}, \bm{\theta})$ is obtained by the mapping $\mathbf{p}=f_\theta(\mathbf{n},\mathbf{c})$ of samples $\mathbf{n} \sim p(\mathbf{n})$.

Specifically, we generate sequential motions in a recurrent manner. Fig.~\ref{fig:norm_flow} shows the details of the generation of motion $\mathbf{p}_i$ at time $i$. The previously generated $k$ motions $\mathbf{p}_{i-k:i-1}$ as well as the audios $\mathbf{a}_{i-k:i+\tau}$ are used as the condition information $\mathbf{c}_i$. Meanwhile, in order to extract the temporal correlations, similar to~\cite{henter2020moglow,alexanderson2020style}, we design the neural network in the normalizing flow layer as a LSTM~\cite{hochreiter1997long} module. Hidden states are updated at each iteration to generate the current motions.
We train this network using a single negative log-likelihood loss. More details about the architecture and the training method can be found in our supplementary materials. 
In the inference phase, one can generate various sequential motions using $\mathbf{p}=f_{\theta}(\mathbf{n}, \mathbf{c})$.
Then these motions can be transferred into realistic videos using the proposed PIRenderer.

\begin{figure*}[t]
\begin{center}
\includegraphics[width=0.95\linewidth]{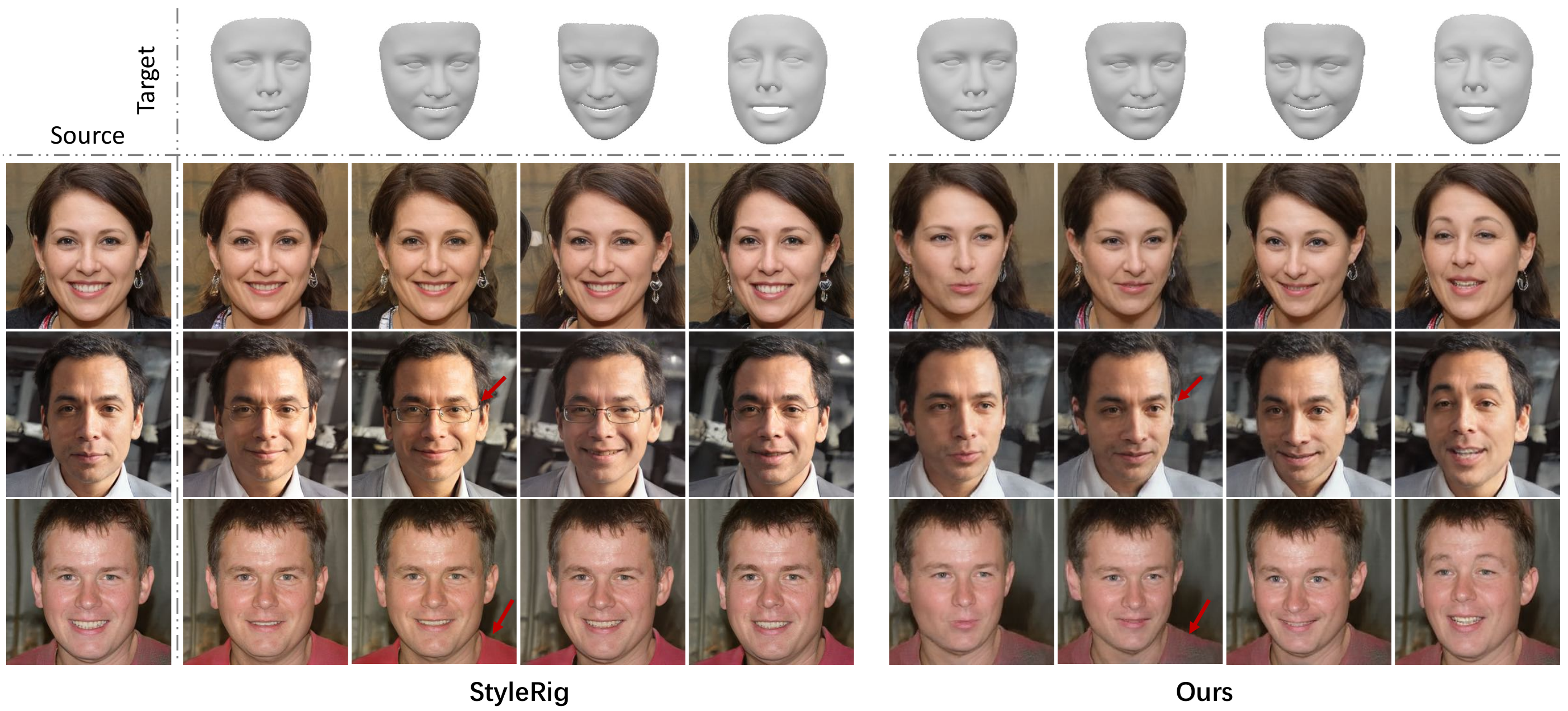}
\end{center}
   \caption{The qualitative comparison results in the intuitive portrait image editing task. The StyleGAN-generated images are used as source images. The left part shows the results of StyleRig~\cite{tewari2020stylerig}. The right part shows the results of our model. We mark the undesired modifications of StyleRig and the corresponding correct results of ours with red arrows.}
 
\label{fig:intuitive_compare}
\end{figure*}

\section{Experiment}

\subsection{Implementation Details}
\label{Implementation_Details}
\noindent
\textbf{Dataset.} We train our model using the VoxCeleb dataset~\cite{nagrani2017voxceleb} which contains 22496 talking-head videos extracted from YouTube videos. We follow the pre-processing method described in paper~\cite{siarohin2019first} to crop faces from the original videos.
The cropped videos contain natural movements where faces move freely within a fixed bounding box. The videos are then resized to $256 \times 256$ for training and testing. 
The corresponding audios are extracted from the videos for the audio-driven reenactment task. 
A total of 17913 training videos and 514 test videos, with lengths varying from 64 to 1024 frames, are obtained.

\noindent
\textbf{Evaluation Metrics.} 
\textit{Learned Perceptual Image Patch Similarity~\cite{zhang2018unreasonable} (LPIPS)} is used to estimate the reconstruction error. 
It computes the perceptual distance between the generated images and reference images.
Meanwhile, the realism of the synthesized results is measured by \textit{Fr\'echet Inception Distance~\cite{heusel2017gans} (FID)} which calculates the Wasserstein-2 distance between distributions of the fake and real images.
The motion accuracy is estimated by the \textit{Average Expression Distance (AED)} and the \textit{Average Pose Distance (APD)} 
which calculate the average 3DMM expression and pose distance between the generated images and targets respectively. 
The trained model of Yu~\etal~\cite{deng2019accurate} is used to extract the 3D face coefficients.
Besides, \textit{Just Noticeable Difference (JND)} tests are performed to evaluate the subjective quality. Volunteers are asked to choose the more realistic one from the data pair of real and fake samples.

\noindent
\textbf{Training Details.} 
We train our model in stages. The mapping network and the warping network are first pre-trained for $200k$ iterations. Then we train the whole model for another $200k$ iteration in an end-to-end manner. We adopt the ADAM optimizer with an initial learning rate as $10^{-4}$. The learning rate is decreased to $2 \times 10^{-5}$ after $300k$ iterations. The batch size is set to $20$ for all experiments.

\subsection{Intuitive Portrait Image Editing}
\label{sec:intuitive_editing}
In this subsection, we evaluate the performance of our model in the intuitive image editing task. 
Although many methods have been proposed for portrait image semantic editing, few of them enable fully disentangled intuitive motion modification. 
In this experiment, we compare the proposed model to a state-of-the-art method StyleRig~\cite{tewari2020stylerig}. 
StyleRig achieves semantic editing by controlling StyleGAN~\cite{karras2019style} via a 3DMM. 
Mapping functions are trained to transform the latent variables of StyleGAN according to the provided target motions described by 3DMM coefficients. 

\begin{table}[]
\centering
\setlength\extrarowheight{1pt}
\resizebox{.38\textwidth}{!}{%
\centering
\begin{tabular}{p{2cm}||P{1.2cm}P{1.2cm}P{1.2cm}}
\hline
         & FID & AED   & APD    \\ \hline
         &  &    &              \\ [-11pt]\hline
StyleRig~\cite{tewari2020stylerig} & \textbf{47.37}    & 0.316 & 0.0919 \\ \hline
Ours     & 65.97    & \textbf{0.257} & \textbf{0.0252} \\ \hline
\end{tabular}%
}
\caption{The evaluation results of the intuitive portrait editing task.}
\label{tab:intuitive_compare}
\end{table}

\begin{figure*}[t]
\begin{center}
\includegraphics[width=0.95\linewidth]{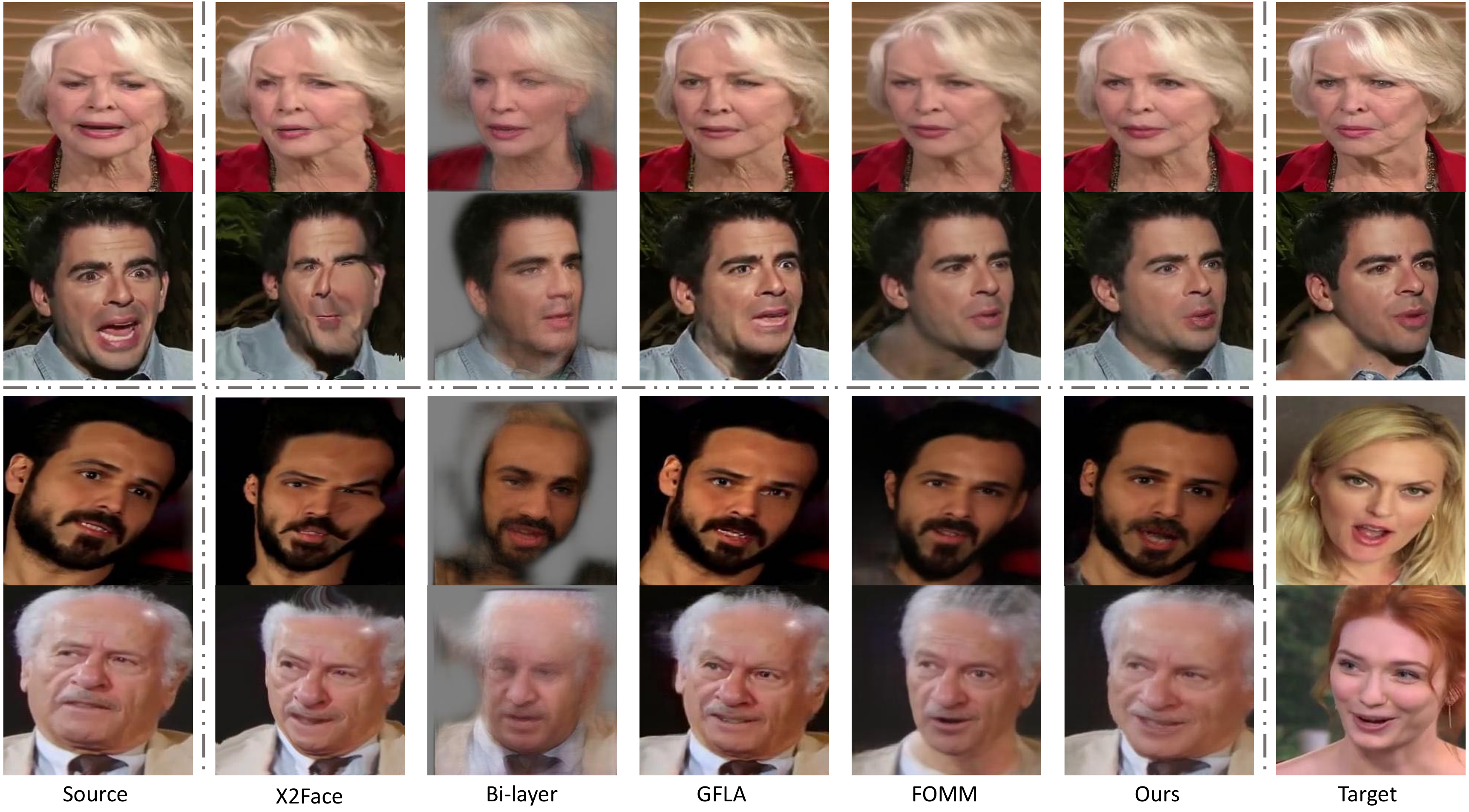}
\end{center}
   \caption{Qualitative comparisons with state-of-the-art methods. The top two lines are the results of the same-identity reconstruction task while the bottom two lines show the results of the cross-identity motion imitation task.}

\label{fig:reenactment_compare}
\end{figure*}

\noindent
\textbf{Editing of StyleGAN-Generated Images.~}
As StyleRig requires the StyleGAN latent variables to perform the modification, we first conduct a comparison with images generated from StyleGAN. 
A total of 500 latent variables are randomly sampled to generate the test images with StyleGAN.
Target motions are obtained by sampling the motion distributions of the FFHQ dataset~\cite{karras2019style}. Similar as~\cite{deng2020disentangled}, several VAEs are trained on the motions extracted from the real images of FFHQ. We train these networks to map 3DMM coefficients to Gaussian distributions and then reconstruct the inputs.
After training, the VAE encoders are discarded and the decoders are used to describe the motion distributions through Gaussian samples.
We randomly sample 100 target motions with various expressions and rotations~\footnote{We do not edit translations here as StyleRig does not support such editing.}, which results in 50$k$ editing images for each model. The evaluation results are shown in Tab.~\ref{tab:intuitive_compare}. It can be seen that by using StyleGAN as the final generator, StyleRig is able to produce more realistic images, which leads to a lower FID score.
However, the higher AED and APD scores indicate that it may fail to faithfully reconstruct the target motions.
Different from StyleRig, our model can generate images with more accurate movements.
Fig.~\ref{fig:intuitive_compare} provides quantitative comparisons. It can be seen that StyleRig generates impressive results with realistic details. However, it tends to generate images with a conservative strategy: motions away from the distribution center are weakened or ignored for better image quality. Meanwhile, some motion-irrelevant factors (\eg glasses, cloth) are changed during the modification.
Although our model is not trained with FFHQ, it can still achieve impressive results when testing with this dataset.
We can generate realistic images with not only correct global poses but also vivid microexpressions such as pouting and eyebrow raising. 
Additionally, the irrelevant source information is well-maintained.

\begin{figure}[t]
\begin{center}
\includegraphics[width=0.95\linewidth]{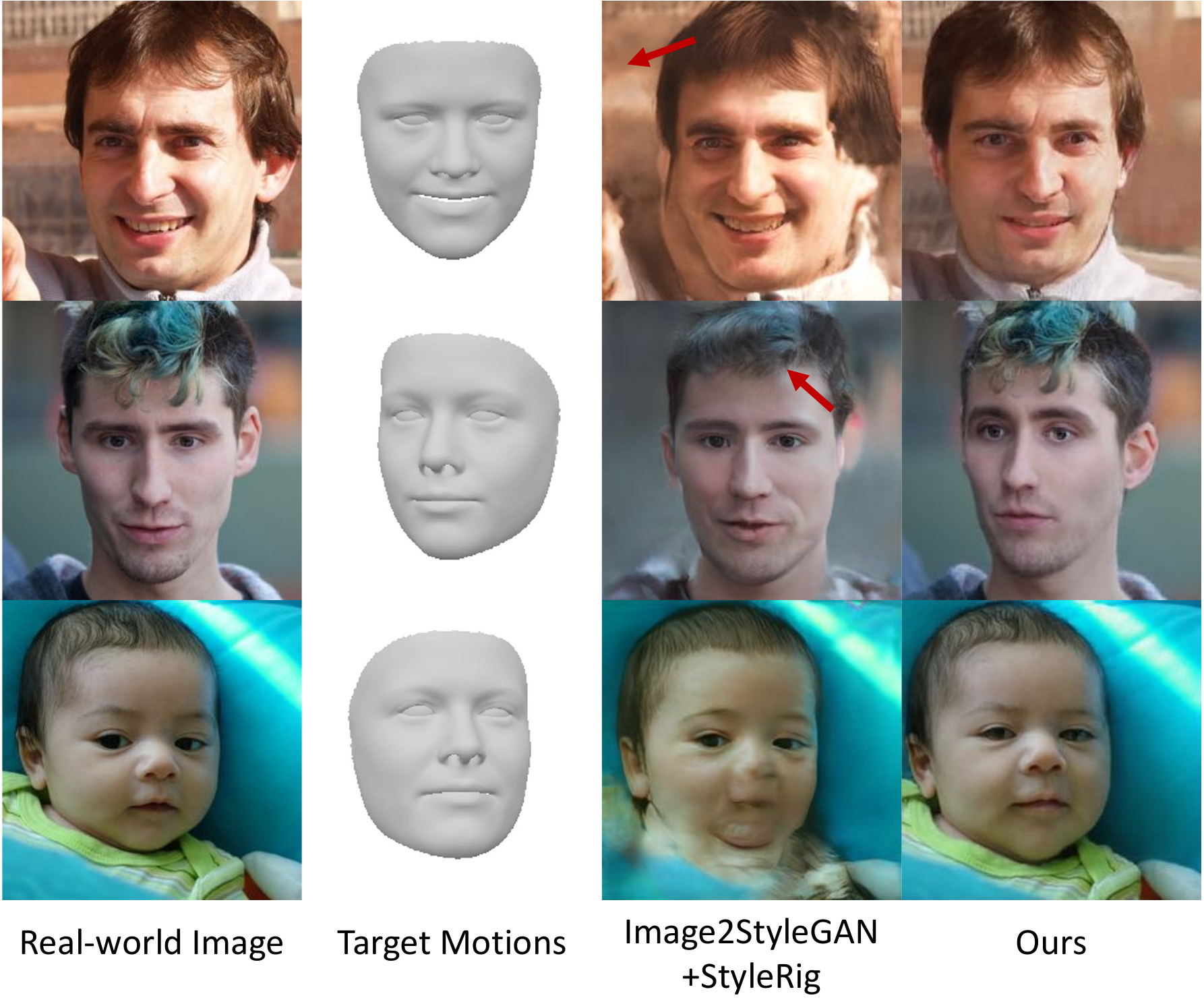}
\end{center}
   \caption{Intuitive editing of the real-world portrait images. Some artifacts are marked by red arrows.}

\label{fig:intuitive_real_image}
\end{figure}

\noindent
\textbf{Editing of Real-World Images.~}
Editing real-world images is a more meaningful task.
However, StyleRig cannot be directly applied for this task, since it transforms portrait images at the StyleGAN feature level.
Fortunately, techniques such as Image2StyleGAN~\cite{abdal2019image2stylegan} can embed real-world images into StyleGAN latent space, which can be employed to help with such editing. The editing results are shown in Fig.~\ref{fig:intuitive_real_image}. 
Since the optimized embeddings produced by Image2StyleGAN may not obey the original latent distributions, the results of StyleRig are often implausible, which is more evident when complex details are shown in the sources.
Our model can be directly applied to edit real-world images.
Meanwhile, we can obtain images with fewer artifacts and realistic source textures.

\begin{table*}[]
\centering
\setlength\extrarowheight{1pt}
\resizebox{0.95\textwidth}{!}{%
\centering
\begin{tabular}{p{2.3cm}||P{1.3cm}P{1.3cm}P{1.3cm}P{1.3cm}P{1.3cm}||P{1.3cm}P{1.3cm}P{1.3cm}P{1.3cm}}
\hline
\multirow{2}{*}{} & \multicolumn{5}{c||}{Same-identity Reconstruction} & \multicolumn{4}{c}{Cross-identity Motion Imitation} \\ \cline{2-10} 
                  & FID        & AED        & APD        & LPIPS   & JND   & FID          & AED           & APD  & JND         \\ \hline
          &       &      &      &      &         &         &   & &      \\      [-11pt]\hline
X2Face~\cite{wiles2018x2face} & 31.75      & 0.2244     & 0.1017     &  0.2400 &  19.0\% &     57.74  &      0.3316     &   0.1319   & 5.4\%   \\ 
Bi-layer~\cite{zakharov2020fast}          & 71.96      & 0.2237     & 0.0449     & 0.2527  & -   & 83.59        & 0.2861        & 0.0494     & -   \\
GFLA~\cite{ren2020deepspatial}              & 12.38      & 0.1555     & 0.0268     & 0.1754   &20.3\%  & 32.17        & 0.2599        & 0.0363   &12.3\%     \\
FOMM~\cite{siarohin2019first}              & 9.114      & 0.1130     & \textbf{0.0161}     & \textbf{0.1220} &34.4\%    & 38.92        & 0.2635        & 0.0336    & 19.4\%    \\
Ours              & \textbf{8.260}      & \textbf{0.1106}     & 0.0164     & 0.1285    &\textbf{36.1\%} & \textbf{28.18}        & \textbf{0.2304}        & \textbf{0.0279} &\textbf{25.5\%}       \\ \hline
\end{tabular}%

}
\caption{Comparisons with state-of-the-art methods on talking-head motion imitation. Both same-identity reconstruction and cross-identity motion imitation are conducted to evaluate the performance. The JND scores of Bi-layer are not provided since the absence of the image background will provide clues to volunteers and cause meaningless evaluation results.}

\label{tab:reenactment_compare}
\end{table*}

\subsection{Talking-head Motion Imitation}
In this subsection, we evaluate the performance of our model in the motions imitation task. 
We compare our model with state-of-the-art methods including: X2Face~\cite{wiles2018x2face}, Bi-layer~\cite{zakharov2020fast}, GFLA~\cite{ren2020deepspatial}, and FOMM~\cite{siarohin2019first}. The released weights of these methods are used for the evaluation if available. The absolute motions are used for all methods. Since Bi-layer does not predict the background, we subtract the background when doing quantitative analyses. 

Two tasks are performed in this experiment: (1) the \textbf{Same-identity Reconstruction} task where the source and driving images are of the same person, and (2) the \textbf{Cross-identity Motion Imitation} task where non-existing videos are generated by mimicking the motions of another individual. 
The evaluation results are summarized in Tab.~\ref{tab:reenactment_compare}. 
It can be seen that FOMM and GFLA achieve impressive results in the same-identity reconstruction task. However, the performance is reduced when driven by cross-identity videos.
One possible explanation is that these methods employ motion descriptors with spatial information (such as sparse keypoints~\cite{siarohin2019first} and facial landmarks~\cite{ren2020deepspatial}), which can benefit the reconstruction task. However, the subject-specific motion descriptors may reduce the performance of the cross-identity reenactment task.
Our model uses a fully disentangled motion descriptor, which helps with achieving better results in the cross-identity reenactment task.
In addition to the objective metrics, a just noticeable difference test is conducted on Amazon Mechanical Turk (MTurk).
Volunteers are asked to choose a more realistic image from image pairs of ground-truth and generated images.
We randomly choose 200 images as the test set for each task. Each pair is compared 5 times by different volunteers.
The fooling rate is shown in Tab.~\ref{tab:reenactment_compare}. It can be seen that our model achieves the best scores in both tasks, which means that our model can generate more realistic results.

The qualitative comparisons are provided in Fig.~\ref{fig:reenactment_compare}. 
It can be seen that the results of X2Face suffer from the warping artifacts because it directly outputs the warped images.
Bi-layer generates images with over-smoothed details, which reduces the realism of the images.
GFLA and FOMM can generate photo-realistic results. However, these methods are sensitive to the occlusions in the driving videos.
Meanwhile, the source identity cannot be well-preserved when the source face shape is very different from that of the target.   
The proposed model generates more realistic results along with accurate movements while still preserving the source identity.

\begin{figure}[t]
\begin{center}
\includegraphics[width=0.90\linewidth]{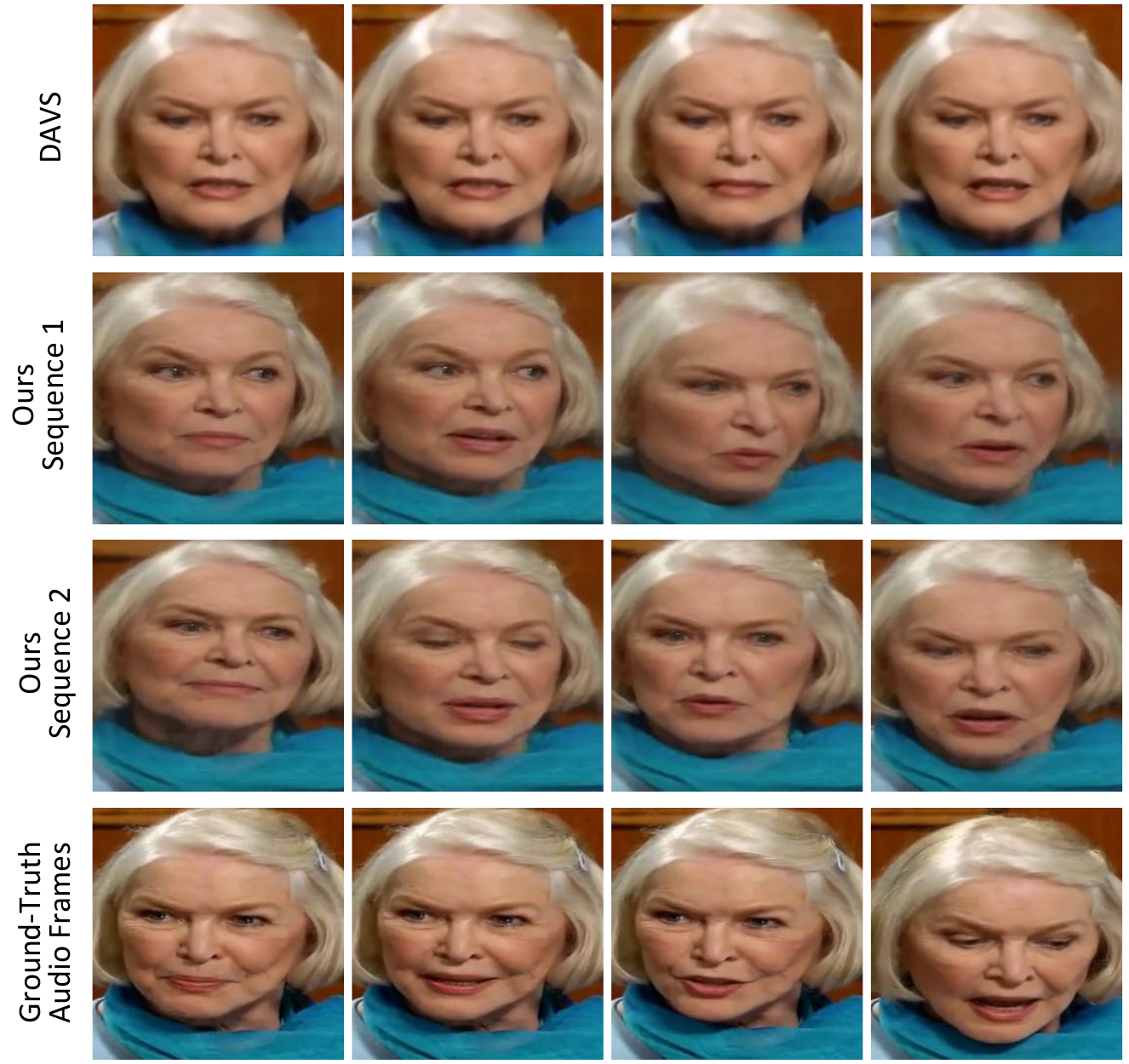}
\end{center}
   \caption{Qualitative results of the audio-driven facial reenactment task.  We provide the results of DAVS~\cite{zhou2019talking} as a comparison. Our model is able to generate various and vivid motions for an audio stream and transfer the motions into realistic videos.}

\label{fig:audio_comparison}
\end{figure}

\subsection{Audio-driven Facial Reenactment}
In this subsection, we show that our model can generate coherent videos by extracting convincing motions from audios. 
Sequential motions with various poses and expressions are first generated by $f_\theta$. Then we transform these motions to arbitrary individuals with our PIRenderer. 
Qualitative results can be found in Fig.~\ref{fig:audio_comparison}.
We provide the results of a state-of-the-art method DAVS~\cite{zhou2019talking} as a comparison to our model.
Evidently, DAVS can extract meaningful lip motions from audios. 
However, it cannot generate motions that are weakly related to audios such as head poses and eye movements.
This inability consequently reduces the realism of the produced videos. 
Meanwhile, DAVS tackles this task with a deterministic network that cannot model the massive variation of the possible motions. 
By employing the semantic coefficients as intermediate results, we can generate various motions from a single audio stream and transfer them into realistic videos. It can be seen that our model not only extracts accurate lip motions but it also generates realistic micro expressions such as pout, blinking.

\section{Conclusion}
We have proposed PIRenderer, an efficient portrait image neural renderer that is able to control faces via semantically meaningful parameters.
Incorporating priors of 3DMMs, our model can conduct intuitive editing of real-world portrait images by modifying the facial expression, head pose, and translation according to user-specified coefficients. Meanwhile, it can also perform motion imitation tasks. Instructed by the subject-agnostic motion descriptors, the model can generate coherent videos with well-maintained source identities. 
We believe that controlling generative neural networks with flexible graphics permits many exciting applications. Extension on the audio-driven facial reenactment task provides an example and shows the potential of such a combination.

\bigbreak

\noindent
\textbf{Acknowledgements.} 
This work was supported by National Key R\&D Program of China (No. 2020AAA0103501). We thank all reviewers for their helpful comments.

{\small
\bibliographystyle{ieee_fullname}
\bibliography{egbib}
}
\clearpage
\onecolumn










\begin{alphasection}
\section{Additional Results of PIRenderer}
In this section, we provide additional results of the proposed model. To achieve more intuitive comparisons, we show the results in a \href{https://youtu.be/gDhcRcPI1JU}{\textit{Supplementary Video}}. In this video, the following materials are provided:
\vspace{-1mm}
\begin{itemize}
  \setlength\itemsep{-0.5mm}
  \item \textbf{Results of the intuitive portrait image editing task}. Although many commercial softwares are available for portrait image editing, complex and high-level modifications (e.g. modifying head posture or expression) are not supported or require professional skills.
  We show that our model can achieve interactive real-world facial image editing, which will greatly reduce the difficulty of image editing and help users to obtain satisfactory images. 

  \item \textbf{Results of the motion imitation task}. In the same-identity reconstruction task, we show that our model can generate coherent videos with realistic details. Meanwhile, compared with the results of GFLA and FOMM, our model is more robust to the occlusions in the driving videos.  In the cross-identity motion imitation task, we show that the proposed model can generate realistic results while preserving the source identity.

  \item \textbf{Results of the audio-driven facial reenactment task}. The generated videos as well as the input audios are provided. It can be seen that our model can generate accurate mouth motions and realistic other motions (eyes, head poses, \etc) for the given audios. Meanwhile, we can generate various motions from only single input audios and transform these motions into coherent videos.

  \item \textbf{Results of the ablation study}. We provide the results of the ablation study described in Sec.~\ref{sec:ablation}. Using the coefficients of a window of continuous frames as the motion descriptors of the center frame can help the proposed model to achieve coherent results.

  \item \textbf{Results of the facial interpolation task}. The results of the facial interpolation task described in Sec.~\ref{sec:interpolation} are reported. It can be seen that our model learns a linear latent-space $\mathcal{Z}$, which enables interpolating images with smooth-varying motions.

\end{itemize}

\section{Analysis of the Target Motion Descriptor}
\label{sec:ablation}
In this paper, we extract 3DMM coefficients using an off-the-shelf 3D face reconstruction model~\cite{deng2019accurate}. 
Although this method produces relatively accurate results, errors and noise are inevitable. 
In our main paper, we mentioned that the estimation errors cause mismatches between the extracted motions and the real motions of target images, which will lead to performance degradation. Meanwhile, failing to model the temporal correlations of videos will cause incoherent videos.
To alleviate these problems, we propose to use the coefficients of a window of continuous frames as the motion descriptors of the center frame.
In this section, we prove the effectiveness of this choice.



An ablation model is trained by using the 3DMM coefficients of a single input frame as the target motion descriptors. 
The evaluation results of the same-identity reconstruction task are shown in Tab.~\ref{tab:ablation}. It can be seen that our PIRenderer can generate images with more accurate target motions. This indicates that our network models the temporal correlations from the coefficients of the continuous frames and thus reduces the errors. Meanwhile, the subjective comparisons are provided in our supplementary videos. It can be seen that the ablation model cannot generate coherent videos, which reduces the reality of the results. Our PIRenderer can generate coherent results with accurate motions.

\begin{table*}[h]
\renewcommand{\thetable}{\Alph{alphasect}.\arabic{table}}

\centering
\setlength\extrarowheight{1pt}
\resizebox{.5\textwidth}{!}{%
\centering
\begin{tabular}{p{2.3cm}||P{1.3cm}P{1.3cm}P{1.3cm}P{1.3cm}P{1.3cm}}\hline
         & FID   & AED    & APD    & LPIPS  \\ \hline
         &       &        &        &        \\      [-11pt]\hline
Ablation & 8.752 & 0.1169 & 0.0182 & 0.1370  \\
Ours     & \textbf{8.260} & \textbf{0.1106} & \textbf{0.0164} & \textbf{0.1285} \\ \hline
\end{tabular}%

}
\caption{Analysis of the target motion descriptor. The ablation model is trained by using the 3DMM coefficients of a single input frame, while our model uses the coefficients of a window of continuous frames as the motion descriptors of the center frame.}
\label{tab:ablation}
\end{table*}

\section{Interpolation of Latent-space $\mathcal{Z}$}
\label{sec:interpolation}
In this paper, we use a mapping network $f_m : \mathcal{P} \rightarrow \mathcal{Z}$ to map motion descriptors $\mathbf{p}$ to latent vectors $\mathbf{z}$. In this section, we show that our model can learn a linear latent-space $\mathcal{Z}$ which supports the task of facial motion interpolation. The interpolated images are generated by latent vectors $\mathbf{z}'$ calculated by: 
\begin{equation}
  \mathbf{z}' = \alpha f_m(\mathbf{p}_1) + (1-\alpha)f_m(\mathbf{p}_2)
\end{equation}
where $\mathbf{p}_1$ and $\mathbf{p}_2$ are two different motions and $f_m$ is our mapping function. 
We first implement the interpolation task with real-world motions. The generated results can be found in Fig.~\ref{fig:interpolation}. It can be seen that our model can generate images with smooth-varying motions. Both expressions and poses are linearly transformed from the motion $\mathbf{p}_1$ to the motion $\mathbf{p}_2$ as the $\alpha$ increasing. 
Then we show that the interpolation can also be performed with a specific motion attribute. Fig.~\ref{fig:interpolation_attr} provides the generated results. 
Given motions $\mathbf{p}_1$ and $\mathbf{p}_2$ with the same expressions but different poses, the interpolated latent vectors $\mathbf{z}'$ control to generate images with the same expressions and smoothly-varying poses. Meanwhile, given motions $\mathbf{p}_1$ and $\mathbf{p}_2$ with the same poses but different expressions, only the facial expressions change in the interpolated results. The facial motion interpolation task can enable many applications such as exemplar-based portrait expression manipulation.  




\begin{figure}[t]
\renewcommand{\thefigure}{\Alph{alphasect}.\arabic{figure}}
\begin{center}
\includegraphics[width=1\linewidth]{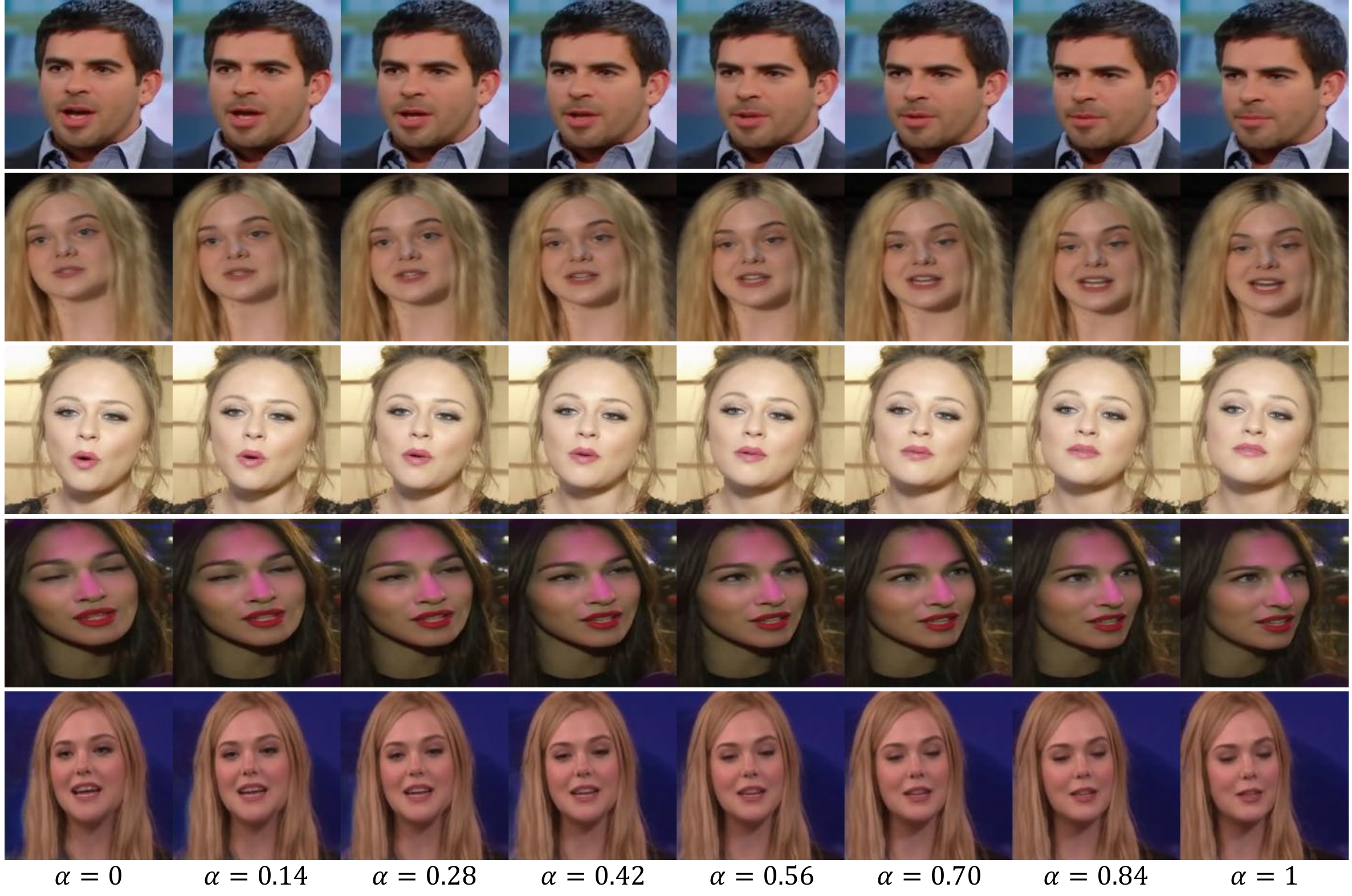}
\end{center}
   \caption{The interpolation results of latent-space $\mathcal{Z}$. Our model can generate images with smooth-varying motions.}
\label{fig:interpolation}
\end{figure}

\begin{figure}[t]
\renewcommand{\thefigure}{\Alph{alphasect}.\arabic{figure}}
\begin{center}
\includegraphics[width=0.99\linewidth]{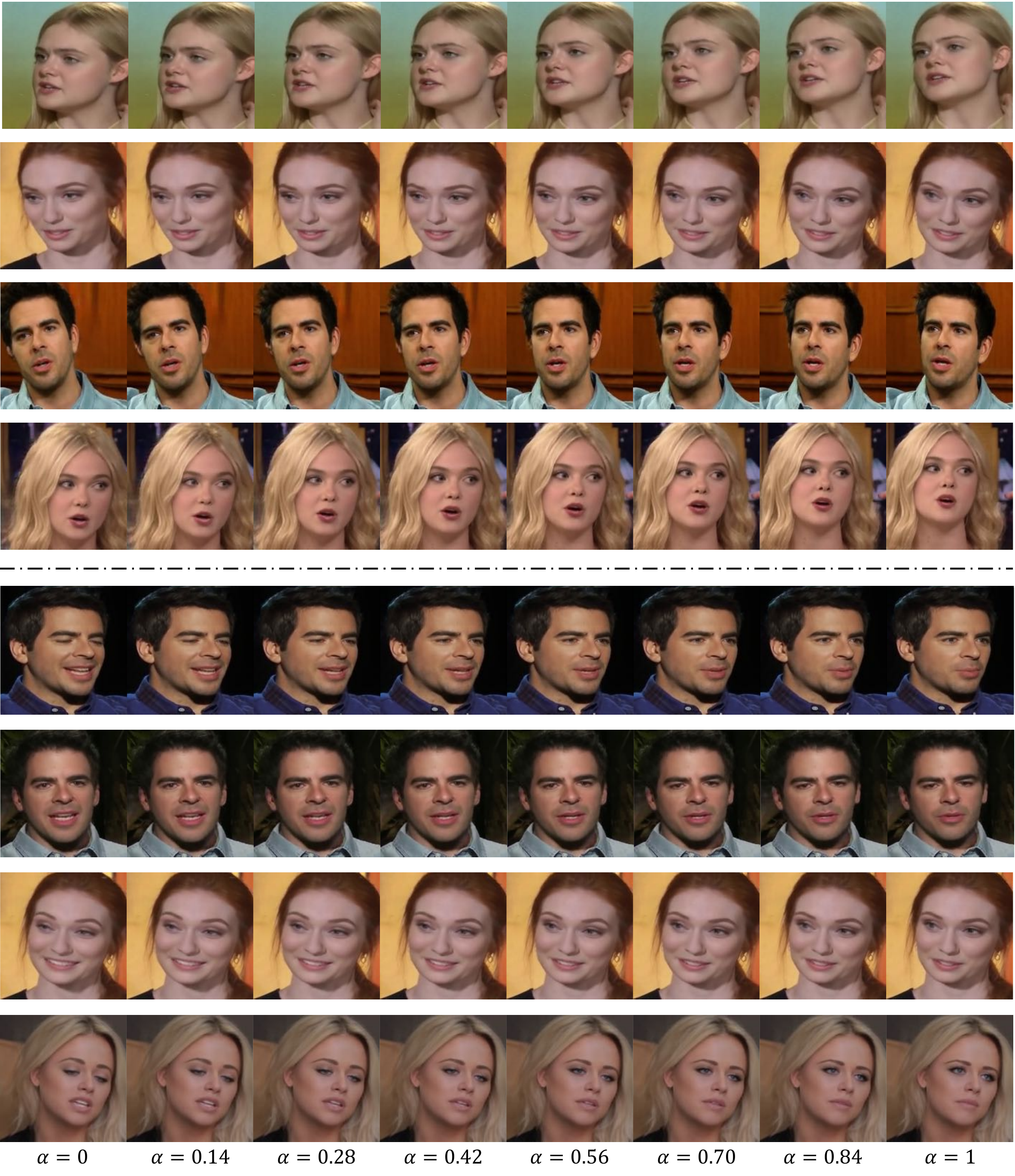}
\end{center}
   \caption{The interpolation results of latent-space $\mathcal{Z}$. The top four rows show the results generated by interpolating motions with the same expressions and different poses. The bottom four rows show the results generated by interpolating motions with the same pose and different expressions.}
\label{fig:interpolation_attr}
\end{figure}

\section{Implementation Details}

\subsection{Implementation Details of PIRenderer}
\textbf{Model Architecture.} The architecture of our PIRenderer is shown in Fig.~\ref{fig:networks}. The mapping network is responsible for transform the target motions $\mathbf{p} \in \mathcal{P}$ into latent vectors $\mathbf{z} \in \mathcal{Z}$. 
As discussed in our main paper, to alleviate the problem brought by the coefficient estimation errors, we use the coefficients of a window of $k$ continuous frames as the motion descriptor of the center frame. We set $k=27$ for all experiments. 
Meanwhile, the 3D face reconstruction models always employ face alignment as the pre-processing method to crop input images such that faces of inputs have similar size and position to improve their performance. Therefore, instead of providing absolute translation parameters $\mathbf{t}$, these methods only estimate relative translations $\mathbf{t}^\prime$. To describe the absolute face positions, we use the cropping parameters $\mathbf{t}_c$ together with the relative translations $\mathbf{t}^\prime$ as our translations. The architecture of the mapping network is shown in Fig.~\ref{fig:networks}. We use 1D convolution layers to process the input motions. Leaky-ReLU is used as the activation function in this network.
The architectures of the warping and editing network are shown in Fig.~\ref{fig:networks}. Auto-encoder structures are used to design these networks.
Skip connections are employed to skip the high-resolution features.
We use the architectures shown in Fig.~\ref{fig:component} as the basic components.
ADAIN operation is used after each convolution layer of (a) ConvDown, (b) ResBlock, and (c) ResBlockUp to inject the latent vectors $\mathbf{z}$. Layer normalization is used as the activation normalization method of the other convolution layers.
We use Leaky-ReLU as the non-linear function in our model.

\textbf{Training and Inference.} We train our model in stages. The mapping network and the warping network are first pre-trained for $200k$ iterations. Then we train the whole model for another $200k$ iteration in an end-to-end manner. We adopt the ADAM optimizer with an initial learning rate as $10^{-4}$. The learning rate is decreased to $2 \times 10^{-5}$ after $300k$ iterations. The batch size is set to $20$ for all experiments. We set $\lambda_w=2.5$, $\lambda_c=4$, and $\lambda_s=1000$. In the inference phase, we use coefficients of $k=27$ continuous frames as the motion descriptors for the reenactment task. In the intuitive image editing task, we repeat the target motion $k=27$ times as the motion descriptors.

\subsection{Implementation Details of $f_\theta$}
We extent our PIRenderer to tackle the audio-driven facial reenactment task by training an additional mapping function $f_\theta$.
The mapping function $f_\theta$ is responsible for generating sequential 3DMM coefficients from audios.
As discussed in the main paper, \textit{Normalizing flow} is employed to design $f_\theta$. The core idea of normalizing flow is to train an invertible and differentiable nonlinear mapping function that maps samples from a simple known distribution (\eg Gaussian) to a more complex distribution. 
In our conditional setting, the function is trained to map motion-condition pairs $(\mathbf{p}, \mathbf{c})$ to latent variables $\mathbf{n}$ with $\mathbf{n} = f_\theta^{-1}(\mathbf{p}, \mathbf{c})$. 
Function $f_\theta$ is composed of a sequence of invertible transformations: $f_\theta = f_1 \circ f_2 \circ \cdots \circ f_K $, such that the relationship between $\mathbf{p}$ and $\mathbf{n}$ can be written as:
\begin{equation}
  \mathbf{n} \xleftrightarrow{f_1(*, \mathbf{c})} \mathbf{h}_1 \xleftrightarrow{f_2(*, \mathbf{c})} \mathbf{h}_2 \cdots \xleftrightarrow{f_K(*, \mathbf{c})} \mathbf{p}
\end{equation}
The key aspect of normalizing flows is that the probability density function $p_{\mathbf{p}|\mathbf{c}}$ can be explicitly computed as:

\begin{equation}
\begin{split}
\log p_{\mathbf{p}|\mathbf{c}}(\mathbf{p} | \mathbf{c}, \theta) &= \log p_{\mathbf{n}}(f_\theta^{-1}(\mathbf{p},\mathbf{c})) +\log \left\lvert \text{det} \frac{\partial f^{-1}_\theta}{\partial \mathbf{p}}(\mathbf{p},\mathbf{c})\right\rvert \\ &= \log p_{\mathbf{n}}(f_\theta^{-1}(\mathbf{p},\mathbf{c})) + \sum_{j=1}^K \log \left\lvert \text{det} \frac{\partial f^{-1}_j}{\partial \mathbf{h}_{j}}(\mathbf{h}_j,\mathbf{c})\right\rvert
\end{split}
\end{equation}
where we define $\mathbf{h}_K \equiv \mathbf{p}$ for conciseness.
We design $f_n$ using a similar architecture as that of Glow~\cite{kingma2018glow}. Each transformation contains three sub-steps: an actnorm function; a linear transformation; and an affine coupling layer; 
Let $\mathbf{x}$  signifies the input of each layer, and $\mathbf{y}$ signifies the output. Both $\mathbf{x}$ and $\mathbf{y}$ are tensors of shape $[c \times t]$ with channel dimension $c$ and time dimension $t$. 
Actnorm is an affine transformation of the activations using a scale and bias parameter per channel with $\mathbf{y}_t = \mathbf{s} \odot \mathbf{x}_t + \mathbf{b}$.
The linear transformation is used to transform the input tensor with the trainable parameters $\mathbf{W} \in \mathbb{R}^{C\times C}$ with $\mathbf{y}_t = \mathbf{W}\mathbf{x}_t$. The affine crouping layer first splits the input tensor $(\mathbf{x}_a, \mathbf{x}_b) = \mathrm{Split}(\mathbf{x})$. Then, a neural network is used to predict the affine parameters $(\log \mathbf{s}, \mathbf{t} ) = \mathrm{NN}(\mathrm{concat}(\mathbf{x}_b, \mathbf{c}))$. Finally the output is obtained by $\mathbf{y} = \mathrm{concat}(\mathbf{s} \odot \mathbf{x}_a + \mathbf{t}, \mathbf{x}_b )$.




Our task requires generating sequential motion descriptors $\mathbf{p}$. Thus, modeling the temporal correlations is a crucial challenge for this task. 
To handle this, we generate the motions in a recurrent manner. The previously generated $k$ motions $\mathbf{p}_{i-k:i-1}$ are used as a part of conditional information for the current generation.
Meanwhile, we design the neural network $\mathrm{NN}(*)$ in the affine coupling layers as LSTM modules to further model the temporal correlations. 
Instead of generating the motions $\mathbf{p}$ using a single $f_\theta$, we design two mapping functions $f_{\theta_1}$ and $f_{\theta_2}$ to generate expressions $\bm{\beta}\in \mathbb{R}^{64}$ and positions $\mathbf{R}\in SO(3),\bm{t}\in \mathbb{R}^3$ respectively. For the expression mapping function $f_{\theta_1}$, the conditional information $\mathbf{c}_i$ consists of two parts: a window of previous expressions $\bm{\beta}_{i-k:i-1}$ and a window of audio single $\mathbf{a}_{i-k:i+\tau}$. 
For the position mapping function $f_{\theta_2}$, in addition to the previously generated motions and the input audios, we further add the initial position of the first frame to the conditional information to help the model building the long-term relationship. We design $K=10$ for $f_{\theta_1}$ and $K=8$ for $f_{\theta_2}$. 
For all experiments, we set $k=5$ and $\tau=6$.

In the training phase, we train our mapping functions to generate latent vectors $\mathbf{n}_i$ from the ground truth motion $\mathbf{p}_i$ and the corresponding conditional information $\mathbf{c}_i$. The negative log-likelihood loss is used as the training loss. 
\begin{equation}
  \mathcal{L}_{nll} = -\log p_n(\mathbf{n}_i) - \sum_{j=1}^K \log \left\lvert \text{det} \frac{\partial f^{-1}_j}{\partial \mathbf{h}_{j}}(\mathbf{h}_j,\mathbf{c}_i)\right\rvert
\end{equation}
In the inference phase, we randomly sample latent vectors $\mathbf{n} \sim p(\mathbf{n})$ and generate the sequential motions using $\mathbf{p} = f_\theta(\mathbf{n}, \mathbf{c})$. The motions in the conditional information $\mathbf{c}_0$ are initialized as the motion of the source image.

\begin{figure}[t]
\renewcommand{\thefigure}{\Alph{alphasect}.\arabic{figure}}
\begin{center}
\includegraphics[width=1\linewidth]{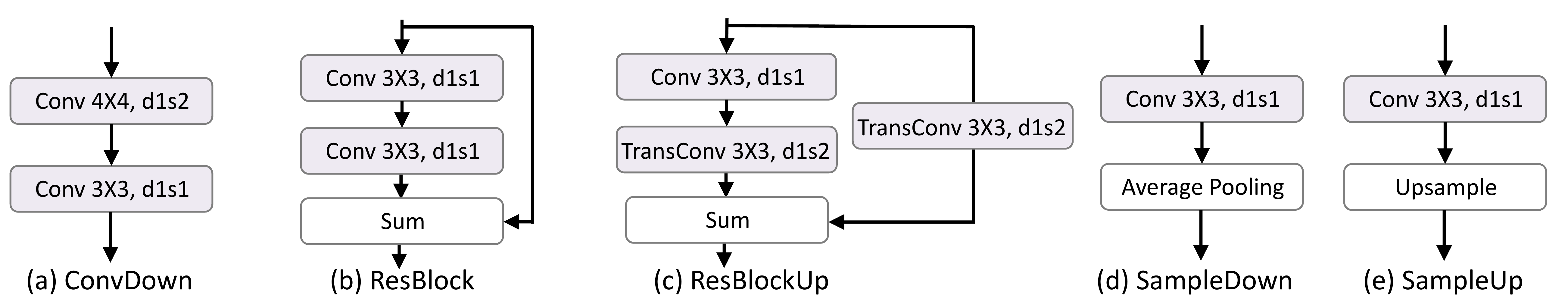}
\end{center}
   \caption{The components used in our networks. The ADAIN operation is used after each convolution layer of (a) ConvDown, (b) ResBlock, and (c) ResBlockUp.}
\label{fig:component}
\end{figure}

\begin{figure}[b]
\renewcommand{\thefigure}{\Alph{alphasect}.\arabic{figure}}
\begin{center}
\includegraphics[width=0.99\linewidth]{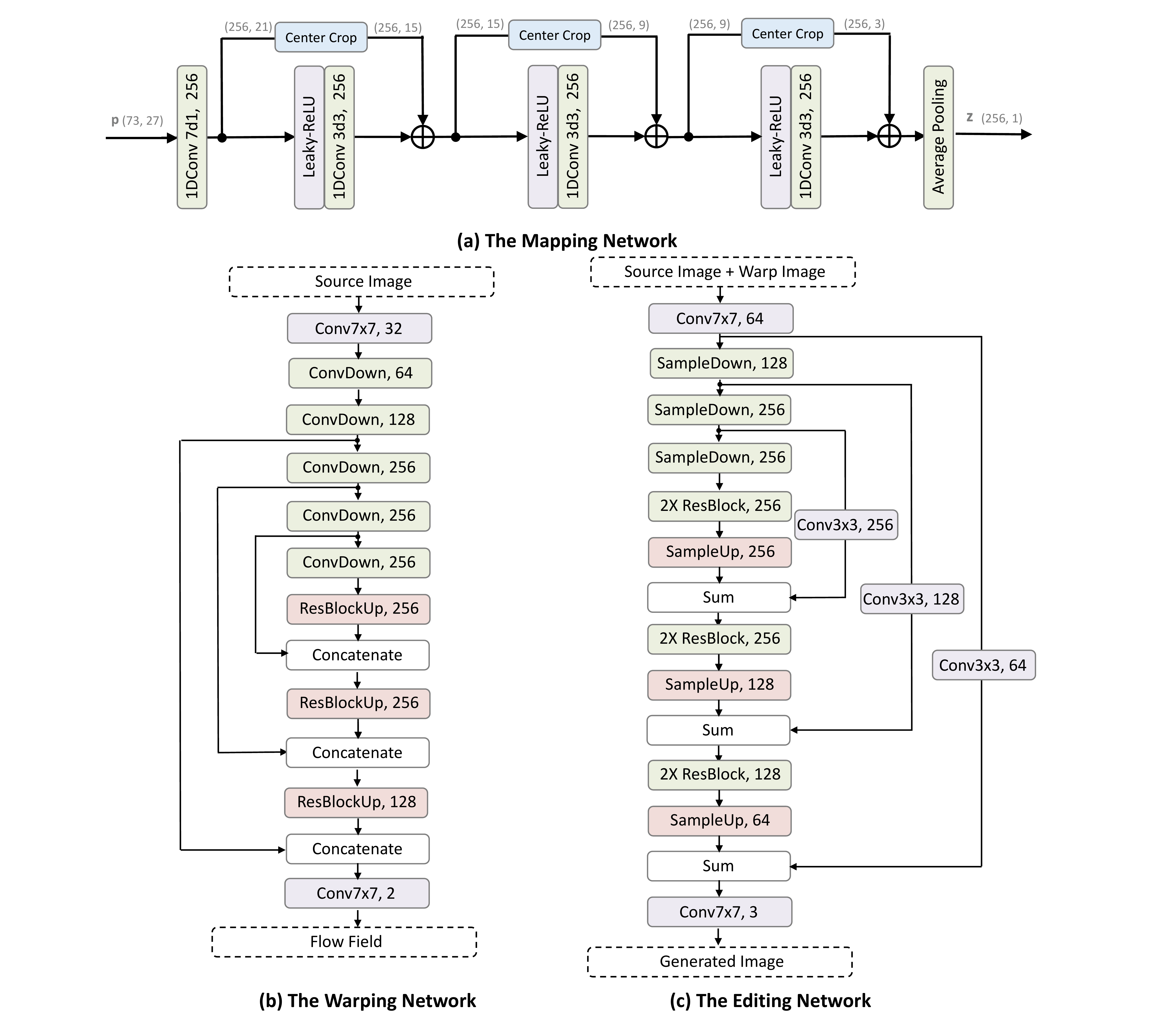}
\end{center}
   \caption{The architecture of our PIRenderer.}
\label{fig:networks}
\end{figure}



\end{alphasection}

\end{document}